# HealthPrompt: A Zero-shot Learning Paradigm for Clinical Natural Language Processing

Sonish Sivarajkumar[1], Yanshan Wang, PhD, FAMIA[1,2,3*]
[1]Intelligent Systems Program, School of Computing and Information, University of Pittsburgh, PA; [2]Department of Biomedical Informatics, University of Pittsburgh, PA; [3]Department of Health Information Management, University of Pittsburgh, PA

**Abstract**

*Deep learning algorithms are dependent on the availability of large-scale annotated clinical text datasets. The lack of such publicly available datasets is the biggest bottleneck for the development of clinical Natural Language Processing(NLP) systems. Zero-Shot Learning(ZSL) refers to the use of deep learning models to classify instances from new classes of which no training data have been seen before. Prompt-based learning is an emerging ZSL technique where we define task-based templates for NLP tasks. We developed a novel prompt-based clinical NLP framework called HealthPrompt and applied the paradigm of prompt-based learning on clinical texts. In this technique, rather than fine-tuning a Pre-trained Language Model(PLM), the task definitions are tuned by defining a prompt template. We performed an in-depth analysis of HealthPrompt on six different PLMs in a no-data setting. Our experiments prove that prompts effectively capture the context of clinical texts and perform remarkably well without any training data.*

**Introduction**

The necessity for vast amounts of annotated text is the fundamental restriction of applying deep neural networks to clinical Natural Language Processing (NLP). Despite some good quality and publicly available clinical corpora, such as i2b2[1] datasets, MIMIC-III[2] datasets, and BioNLP[3] datasets, more annotated clinical text datasets from real-world Electronic Health Records (EHRs) are needed to build clinical NLP systems. Most of the existing machine learning-based clinical NLP systems are based on the supervised approach, requiring abundant data for reasonable accuracy. Supervised learning has been used in clinical NLP for different applications, including disease identification[4], cohort selection[5], drug regimen selection[6], etc. These systems have proved to be accurate and effective in scenarios with abundant training data availability. However, the reality is that most clinical NLP projects have only a small number of annotated clinical documents due to the expensive and time-consuming manual annotation process. Therefore, building clinical NLP systems with few or no annotated documents remains a top focus in informatics research.

A Pre-trained Language Model (PLM)[7] is a neural network trained on a vast amount of unannotated data (such as Wikipedia data or PubMed data) in an unsupervised way. This process is called pre-training, which allows the model to learn the general structure of a language (or domain), including the usage of vocabulary in that domain-specific context. The state-of-the-art biomedical and clinical PLMs like *BioBERT*[8] and *Clinical BERT*[9] have been trained using millions of clinical texts, including EHRs and radiology reports. The model is then transferred for a downstream NLP task, where a smaller task-specific labelled dataset is used to tune the PLM and thus construct the final model capable of executing the downstream task. This process is called fine-tuning a PLM. The entire paradigm is called transfer learning, as the model learns a general context during pre-tuning, and this knowledge is transferred for a specific NLP task by fine-tuning the model in a low-data regime. For example, *Clinical BERT* has been pre-trained largely on unlabeled clinical text datasets like MIMIC, during which it has learned clinical linguistic characteristics from clinical narratives such as physician's notes. Later, this general clinical knowledge could be transferred for specific tasks like adverse event detection[10] or clinical Named Entity Recognition (NER)[11] tasks, by finetuning the PLM with a lesser amount of task-specific annotated data.

The approach described above with few annotated documents is usually called few-shot learning (FSL). While in real-world scenarios, there may not be any annotated data for a medical concept. Zero-shot learning (ZSL)[12] is emerging research in NLP which doesn't require any laborious data annotation by experts. In ZSL, models pretrained from unlabeled data can be used for making predictions on unseen data. This technique gained popularity after PLMs have been successfully applied on cross-domain adaptation tasks[13]. Researchers have proved that PLMs can be effectively used for sentiment analysis tasks in a zero-shot cross-lingual setting[14]. In clinical NLP, there are numerous medical

---

concepts that need to be extracted from clinical notes and meanwhile, very few annotated datasets are available. However, ZSL has been rarely investigated in clinical NLP. Therefore, there is an urgent need to investigate the use of ZSL systems in clinical NLP, which doesn't require annotated data for downstream applications.

In this paper, we investigate the use of the ZSL technique in addressing the issue of the lack of annotated datasets for clinical NLP tasks. We prove that clinical text classification can be executed in a no-data setting by tuning a PLM to predict the probability of classifying a text. We used prompt-based learning as the ZSL approach. Prompt-based learning[15] is one of the latest zero-shot learning methods, which involves creating templates for a given NLP task. This process of building prompting functions or templates for the input text data is called prompt engineering. We can define a mapping function that maps this set of tokens to the expected prediction for the downstream task. In short, prompt-based learning can be defined as a two-step process, where we apply a template with input and output [Z] slots and use the knowledge learned by a pre-trained language model to predict the token that can be filled in the slot [Z].

An example of the prompt-based learning process is illustrated in Table 1. We make the PLMs to fill in the masked tokens in sequences like *"The patient has cough and expanded chest that does not deflate when he exhales. This is the symptom of [MASK] disorder"*. Then the PLM constructs an explicit word-context or word co-occurrence matrix from the pre-trained embeddings. Using prompt-based learning, we try to tune the PLMs to make predictions by defining these sequences or prompts. Till the date of writing this paper, prompts have not been used for NLP systems in the clinical domain to the best of our knowledge.

**Table 1**. An example of the prompt-based learning process. We define prompting functions or templates, which transform the original input text into forms that can be filled by a PLM that predicts the probabilities of texts that can be served in the slot.

| Description | Input |
| --- | --- |
| Input text which needs to be classified | "The patient has cough and expanded chest that does not deflate when he exhales." |
| We can define a Prompting Function | [text]. This is a symptom of [Z] disorder. |
| Now the transformed prompt becomes | "The patient has a cough and expanded chest that does not deflate when he exhales. This is a symptom of [Z] disorder" |
| The language models predict a token, phrase or sentence that fills the slot [Z] | "lung", "asthma", or "respiratory". |

**Background**

With the development of PLMs, in other words Neural Language Models, various architectures have been proposed to solve NLP tasks. The first-generation PLMs like Skip-Gram[16] and GloVe[17] were based on word embeddings. A major drawback of these models was that they are not capable of understanding complex linguistic concepts and the underlying contexts behind these embeddings. Nevertheless, they were effective in comprehending the semantic meaning of a word/sentence.

The second-generation PLMs are based on contextual embeddings. These models can differentiate the semantic meaning of words in different contexts. *BERT*[18], *GPT-2*[19], *RoBERTa*[20], *XLNet*[21], *T5*[22] are examples of the second-generation PLMs that can be fine-tuned for various NLP tasks. These models have achieved state-of-the-art performance for various NLP tasks. Table 2 is a summary of the six PLMs used in this study.

Prompt-based learning is one of the latest developments in the field of Natural Language Processing and has not been effectively explored for many Natural Language Inference (NLI) applications. There have been multiple attempts to use pre-trained language models are unsupervised learners.

Radford et al.[23] demonstrated that pretrained language models could be used for many NLP tasks by "task descriptions", without the need for fine-tuning They have used this method for reading comprehension, translation, summarization, question answering tasks, etc. and found that GPT-2 could produce good results in most of these tasks without any explicit supervision.

Researchers have also tried to use semi-supervised training procedures with cloze-style templates for Text classification and NLI tasks[24]. This few-shot approach outperformed supervised learning in a low-resource setting by a large margin. GPT-3[25], with its 125-175 Billion parameters, had achieved good results in the zero-shot and one-shot settings and occasionally outperformed state-of-the-art models in the few-shot scenario[25]. During the LAMBADA dataset test, GPT-3 shows 86.4% accuracy in the few-shot settings, an increase of 18% over previous state-of-the-art (SOTA) models.

**Table 2**. A summary of PLMs used in this study.

| PLM | Description |
| --- | --- |
| BERT[18] | Bidirectional Encoder Representations from Transformers(*BERT*) is a pre-trained language model developed by Google, which was trained using 800 million words from the BookCorpus dataset and 2.5 billion words from Wikipedia. It is based on Transformer architecture[26], a neural network architecture for language interpretation built on a self-attention mechanism. |
| RoBERTa[20] | Robustly Optimized BERT Pre-training Approach(*RoBERTa*) is based on *BERT* architecture. It is trained with a bigger batch size and longer text sequences without Next Sentence Prediction(NSP). It contains more parameters than BERT (123 million for RoBERTa base and 354 million for RoBERTa large). It is trained using dynamic masking patterns, in which the masked token is altered with each iteration of the input. |
| BioBERT[8] | Bidirectional Encoder Representations from Transformers for Biomedical Text Mining (*BioBERT*) has the same architecture as BERT but is mainly pre-trained on Biomedical texts. It is trained on PubMed abstracts (PubMed) and PubMed Central full-text articles (PMC), apart from the Wikipedia and BookCourpus data. |
| Clinical BERT[9] | This BERT model is primarily trained on clinical texts, EHRs and discharge summaries and thus produces better performance than other models on EHR text corpus. |
| GPT-2[19] | This is essentially a decoder-only transformer. The model is built by stacking up the transformer decoder blocks. Unlike the self-attention used by transformers, GPT-2 uses masked self-attention. It is an autoregressive language model. |
| T5[22] | Text-to-Text Transfer Transformer (*T5*) is one of the latest PLMs released by Google, which outputs a text string instead of a label or a span of the input to the input sentence. Thus, it is a seq2seq model. Instead of employing task-specific architecture for different sorts of language tasks, it was built to have one standard architecture to learn many tasks |

Based on unlabeled corpora and a pre-defined label schema, researchers have shown that the prompt-learning approach of Zero-shot learning can automatically learn and summarize entity kinds[27]. This approach could outperform even 32-shot fine-tuning. They have also shown that text classification tasks can be formulated as a prompt-based learning problem, given by:

$$p(y \in \mathcal{Y} \mid x) = p([\text{MASK}] = w \in \mathcal{V}_y \mid T(x))$$

where $y \in \mathcal{Y}$ is a label, and $\mathcal{V}_y = [l_1, l_2,..l_n]$ is the label word set with labels $l_1, l_2,..l_n$. A prompt template $T$ modifies the input text $x$ and creates a prompt function $T(x)$, with additional tokens than $x$. [MASK] can be considered the slot here, which is predicted by the pre-trained language model from the label set $\mathcal{V}$.

**Methods**

*HealthPrompt Paradigm*

We propose a novel clinical NLP framework using the prompt-based zero-shot learning named HealthPrompt. The architecture of the HealthPrompt framework is shown in Figure 1. HealthPrompt comprises the following components:

1. EHR Chunk Encoder: Each EHR document is split into segments of texts or chunks. This forms a chunk-level representation of an EHR. Details will be discussed in the following sections.
2. Label Definition: A label set is created, containing the labels to be predicted by the zero-shot clinical text classification model.
3. PLM Selection: Choosing a PLM to predict the prompt class. Different PLMs have different properties, and hence, they perform differently in each task. E.g., *BioBERT* works better on Biomedical data, while *Clinical BERT* works best on clinical data.
4. Template Definition: The most crucial part of the framework. Prompt templates are the text templates that modify the input text before being fed into a PLM. These templates need to be carefully engineered for the downstream task.
5. Inference: PLM infers the most appropriate label that fits the prompt template. This is done based on an answer search technique, which is discussed in the following sections.

The key feature of the HealthPrompt NLP framework is that it does not require any training data. We can directly feed the unlabeled EHR corpus to the framework after defining the prompt template and label set. Thus, HealthPrompt is a ZSL framework based on prompt-based NLP on Clinical texts. The input EHR document is first split into chunks of sentences. These chunks or segments of sentences are then passed into the prompt-based model, which infers the label of the clinical texts.

HealthPrompt can be used for any clinical NLP task that can be formulated as a classification problem, including clinical Named Entity Recognition (NER), clinical text classification, and Adverse Drug Event (ADE) detection. In our implementation, the Openprompt[24] python library was used to build a prompt-based zero-shot model. This unified python library gives the flexibility to define the prompt templates and verbalizers (i.e., label sets).

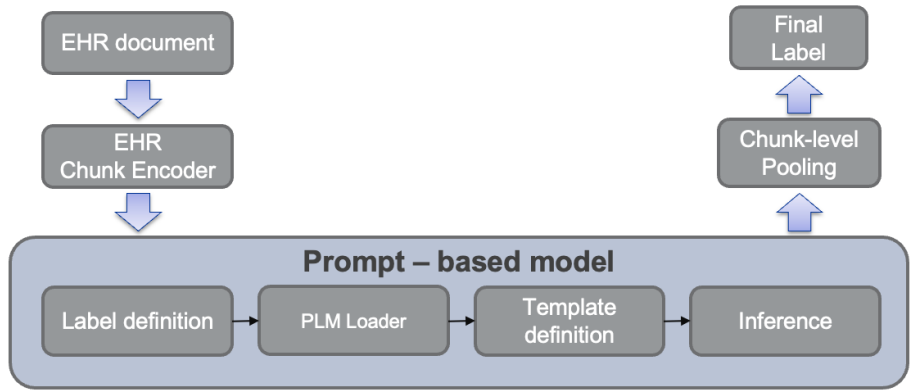

**Figure 1**. The architecture of the HealthPrompt framework.

*Chunk Encoder and Pooling*

Most PLMs are limited by the maximum length of input text due to hardware limitations. BERT, for instance, supports only 512 tokens at a time. But EHR documents usually contain thousands of tokens. For instance, each document in MIMIC-III contains 8,131 tokens on average.

Chunk-level pooling[28] is a solution to this problem. In this technique, Chunk Encoder splits each EHR document into chunks(segments) of equal lengths. The hierarchical encoding and pooling framework was originally designed to overcome the limitation of the PLMs to be trained on long documents. We adopted this chunk representation in the HealthPrompt framework, such that each chunk is passed to the prompt-based model, and the predicted label of all the chunks of a document are collected as a label set or collection set of the document, as illustrated in Figure 2. Chunk Pooler aggregates the predicted labels from the label collection. We use maximum pooling to combine the chunk labels into document labels, and thus perform a document-level classification of long clinical text documents. Short documents or clinical texts can also be classified using HealthPrompt, where they will be split into a smaller number of chunks. HealthPrompt is flexible for sentence-level clinical text classification, where the chunk encoder considers the sentence as one chunk, and hence no splitting is performed on the clinical text.

The labels corresponding to each chunk of the document is then normalized with respect to their number of occurrences in the label collection set of the document, and the label with maximum probability is chosen as the final label of the entire document. Let L be the label set with *L1, L2,.., Ln* labels corresponding to chunk *C1, C2, ..., Cn*. The final label *Lf* is given by:

$$Lf = Max[P(Ln/L)]$$

The type of encoder and the pooling mechanism used by HealthPrompt determines how a document is represented. Though we have used maximum pooling, HealthPrompt gives the flexibility to use other pooling mechanisms like average pooling, transformer encoder, Long Short-Term Memory (LSTM), and Convoluational Neural Network (CNN) based pooling.

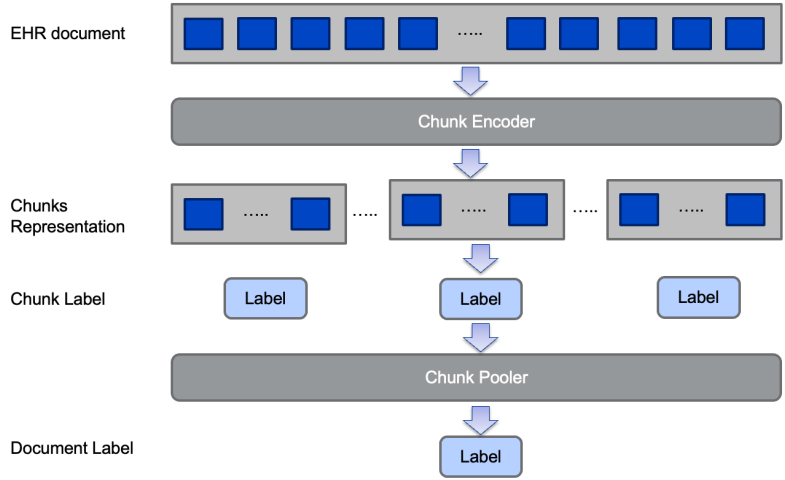

**Figure 2**. The architecture of Chunk-level pooling in HealthPrompt.

### *Template Definition: A Prompt Engineering Step*

Prompt Engineering is related to the Template Definition of the HealthPrompt NLP framework. Different templates can be used to express the same concept. Hence it is essential to carefully design the templates for exploiting the zero-shot capability of a language model.

*Cloze Prompt*. Cloze prompts are templates with slots in a continuous textual string. This can be viewed as a fill-in-the-blank query on the LM. Cloze prompts have been effective in an open-domain question answering system[29]. Researchers have also proved that cloze templates using BART achieve competitive results on the NER benchmark and outperform traditional sequence labelling methods[30]. In general, Cloze prompts are used for text classification and NLU (Natural Language Understanding) tasks. "[X]. This is [Z] disease." is an example of cloze prompts.

*Prefix Prompt*. In prefix prompts[31], tokens predicted by a PLM fill the subsequent masks. These are commonly used for text generation and sequence prediction tasks. Li et al.[32] experimented with prefix tuning and showed that these prompts could improve generalization on Natural Language Generation (NLG) tasks. "[X]. Disease: [Z]" Is an example of prefix prompts.

In this study, we created four different prompts for a clinical text classification task, whereby each patient note is classified into different disease labels using prompt-based learning without training the model. Among these prompts, two were cloze prompts, and two were prefix prompts, which are given in Table 3.

**Table 3**. Four different prompt templates used for the experiment.

| Prompt Template | Type |
|---|---|
| '{"text"}. Disease : {"mask"}' | Prefix Prompt |
| '{"text"} : This effects {"mask"}' | Prefix Prompt |
| '{"text"} : {"mask"} disorder' | Cloze Prompt |
| '{"text"} : {"mask"} type of disease' | Cloze Prompt |

*Pretrained Language Models (PLM)*

The proposed HealthPrompt Clinical NLP framework could embed any PLM. In this study, we experimented with six PLMs, namely *BERT, BioBERT, Clinical BERT, RoBERTa, GPT-2* and *T5*. We used these language models for the clinical text classification tasks, and evaluated the four prompt templates using the six PLMs considered in this study. We would like to validate whether careful prompt engineering is effective for clinical NLP tasks and which PLM has the best performance in the HealthPrompt framework.

*Inference using Answer search*

A prompt task is defined by a label set and by prompt engineering. A label set is a list of labels on which the input text has to be classified. Once the label set and prompt template are defined, PLMs generate multiple tokens that can fill a prompt template. For instance, "Patient has severe headache and nausea. This is a symptom of [MASK]". In this case, a PLM can fill the [MASK] with tokens like malaria, brain injury, brain aneurism, etc.

Answer search is the process by which a PLM selects the token which fits best for a task definition. It is performed over the set of potential answers based on the probability of each value in completing the prompt. A verbalizer[33] function is defined for the PLM to predict the probability distribution over the label set. This set of logits is then mapped to the corresponding labels. The label with the highest probability is selected as the predicted token.

**Dataset**

We used the phenotype annotation dataset for patient notes in the MIMIC-III database[34] as the testing dataset. This dataset identifies whether a patient has a given phenotype based on their patient note. The patient notes were retrieved from MIMIC-III, a dataset collected from Intensive Care Units of a large tertiary care hospital in Boston. Those notes were manually annotated for the presence of several high-context phenotypes relevant to treatment and risk of re-hospitalization. We randomly sampled a subset of 347 patient notes and the corresponding phenotype annotations to evaluate the performance of our zero-shot HealthPrompt framework. These long patient notes had around 7651 tokens on average. This subset consisted of 10 classes of different diseases. The distribution of the dataset and its labels is shown in Table 4.

These documents are split by the Chunk-level enoding mechanism of the HealthPrompt to get chunk level labels, which is aggregated to by the pooling mechanism to fetch the final document label.

**Table 4.** Distribution of different phenotype categories in the MIMIC-III phynotype subset used in this study.

| Phenotype Category | Number of documents |
|---|---|
| Chronic Neurological Dystrophies | 54 |
| Advanced Heart Disease | 49 |
| Depression | 41 |
| Advanced Cancer | 38 |
| Advanced Lung Disease | 38 |
| Chronic Pain Fibromyalgia | 35 |
| Obesity | 33 |
| Non-Adherence | 24 |
| Alchohol Abuse | 24 |
| Dementia | 11 |

**Evaluation**

We used the entire dataset with 347 patient notes as the testing set for the HealthPrompt clinical NLP framework. All four prompts were evaluated on the testing set, with each of the six PLMs we considered for this study. Accuracy, precision, recall, and F1 score were calculated for each experiment. The following are the definitions of these metrics:

$$Accuracy = \frac{True\ Positive + True\ Negative}{True\ Positive + True\ Negative + False\ Positive + False\ Nagative}$$

$$Precision = \frac{True\ Positive}{True\ Positive + False\ Positive}$$

$$Recall = \frac{True\ Positive}{True\ Positive + False\ Nagative}$$

$$F1\ score = \frac{2 \cdot True\ Positive}{2 \cdot True\ Positive + False\ Positive + False\ Nagative}$$

**Results**

We report all models' performance in Table 5 for four prompt templates. For the MIMIC-III phenotype dataset, we report the average accuracy, precision, recall, and Macro F1 score across the ten classification labels. Cloze prompt "[X]. [MASK] type of disease" was the best performing prompt for the phenotype classification task with the best F1 score of 0.86 and an accuracy of 0.85. The prefix prompt was also able to perform well, with the best F1 score of 0.81 and accuracy of 0.81. The other two prompts received comparatively low performance with all the models, with accuracy scores in the range of 0.50-0.75. Hence, it is clear that cloze prompts produce results comparable to the state-of-the-art clinical text classification models by tuning them using a prompt definition.

**Table 5**. Results on the MIMIC-III phenotype subdataset with six PLMs.

| Prompt Template | PLM Model | Accuracy | Precision | Recall | F1-score |
|---|---|---|---|---|---|
| '{"text"}. Disease : {"mask"}' | BERT | 0.61 | 0.68 | 0.69 | 0.68 |
| | BioBERT | 0.78 | 0.78 | 0.71 | 0.74 |
| | **Clinical BERT** | **0.81** | **0.79** | **0.83** | **0.81** |
| | RoBERTa | 0.69 | 0.71 | 0.71 | 0.71 |
| | GPT-2 | 0.68 | 0.67 | 0.57 | 0.62 |
| | T5 | 0.63 | 0.64 | 0.66 | 0.65 |
| '{"text"} : This effects {"mask"}' | BERT | 0.55 | 0.45 | 0.41 | 0.43 |
| | BioBERT | 0.71 | 0.70 | 0.71 | 0.70 |
| | **Clinical BERT** | **0.73** | **0.72** | **0.72** | **0.72** |
| | RoBERTa | 0.61 | 0.60 | 0.57 | 0.58 |
| | GPT-2 | 0.51 | 0.50 | 0.61 | 0.55 |
| | T5 | 0.65 | 0.69 | 0.61 | 0.65 |
| '{"text"} : {"mask"} disorder' | BERT | 0.61 | 0.68 | 0.69 | 0.68 |
| | BioBERT | 0.71 | 0.69 | 0.70 | 0.69 |
| | **Clinical BERT** | **0.75** | **0.73** | **0.72** | **0.72** |
| | RoBERTa | 0.58 | 0.51 | 0.48 | 0.49 |
| | GPT-2 | 0.69 | 0.71 | 0.71 | 0.71 |
| | T5 | 0.68 | 0.69 | 0.70 | 0.69 |
| '{"text"} : {"mask"} type of disease' | BERT | 0.70 | 0.74 | 0.73 | 0.73 |
| | BioBERT | 0.77 | 0.73 | 0.73 | 0.73 |
| | **Clinical BERT** | **0.85** | **0.86** | **0.86** | **0.86** |
| | RoBERTa | 0.71 | 0.71 | 0.69 | 0.70 |
| | GPT-2 | 0.73 | 0.73 | 0.75 | 0.74 |
| | T5 | 0.71 | 0.71 | 0.72 | 0.71 |

*Clinical BERT* showed the best performance in all evaluation metrics in all the prompts, including F1-score and accuracy. *BioBERT* also produced comparable results to *Clinical BERT*. Despite having a complex model and significantly a greater number of parameters, GPT-2 and T5 showed poor performance compared to others. The potential reason is that *GPT-2* and *T5* models are autoregressive and seq2seq models, respectively. Our prompt approach mainly relied on masking a token in the prompt template. Hence, Masked Language Models (MLM) performed better than *GPT-2* and *T5*. Within Masked Language Models, *Clinical BERT* and *BioBERT* performed better because these models were explicitly trained on Biomedical and Clinical literature.

These results show that prompt-based learning can effectively apply the *Clinical BERT* model to clinical text classification tasks in a no training data setting. Using HealthPrompt, we directly utilized the pre-trained language models and tuned the task definitions with prompts. This alignment of clinical task definition to pre-trained language models produced significantly high performance, on par with any of the text classification models in the general domain.

**Discussion**

Recently prompt-based ZSL has become a hot research area in NLP. Most existing prompt-based NLP research focused on simple sentence-level classification tasks. However, clinical texts from EHRs are mostly long text documents that differ from these sentence-level data in the general domain. This is one of the biggest challenges when applying prompt-based ZSL and PLM to clinical NLP. In the HealthPrompt framework, we incorporated chunk-level encoding and pooling to tackle this problem.

HealthPrompt can serve as a general NLP framework for any clinical NLP tasks, when we do not have any training data at the initial stages of a project. Particularly, at the beginning of the outbreak of COVID-19 pandemic, the HealthPrompt could be rapidly applied to classify new disease and phenotypes based on the clinical texts. In addition, most of the clinical text datasets are not publicly available and may require many agreements and attestations to access the data, which is one of the biggest impediments in Clinical NLP research. The proposed HealthPrompt framework can be effectively used in most of the clinical NLP tasks by carefully designing different prompt templates, regardless of the data availability and NLP expertise.

**Limitations and Future Work**

There are three major limitations in this study. First, only six PLMs were tested in the HealthPrompt framework. Among the six PLMs, four were general domain PLMs, namely *BERT, RoBERTa, GPT-2,* and *T5*, and two were clinical domain PLMs, namely *BioBERT* and *Clinical BERT*. This was due to the lack of availability of clinical domain-specific PLMs. Nevertheless, we were able to produce results close to SOTA clinical NLP systems with *Clinical BERT* and *BioBERT*, without using any training data. Specific target-domain PLMs have been developed for domain-specific NLP tasks. For example, *COVID-Twitter-BERT*[35] is a such attempt, where a transformer-based PLM was developed by pretraining on a vast corpus of COVID-19 related tweets. Researchers have also released fine-tuned PLMs that can be used for specific applications, as *BioBERT* finetuned on COVID-19 datasets. These finetuned domain-specific PLMs could produce better results on the HealthPrompt framework for the corresponding downstream task, which will be tested in our future work.

Second, only four prompt templates were designed and utilized in the HealthPrompt framework. The reason is that testing different prompt templates is out of scope of this study as our primary goal is to develop a prompt-based ZSL clinical NLP framework. More evaluations with the new prompt templates and with the development of new domain-specific PLMs are subject to a future work.

Third, the testing dataset used in this paper is relatively small. Only 347 clinical documents from the MIMIC-III phnotype dataset was used to test the HealthPrompt framework. The reason is that most documents are lengthy documents with hundreds of sentences and the chunk encoding and pooling component in HealthPrompt is computationally expensive to process these document. In the future work, we would like to investigate more efficient way of processing lengthy documents and test HealthPrompt on larger datasets.

**Conclusion**

The lack of such publicly available datasets is the biggest bottleneck for the development and widespread adoption of deep learning techniques in clinical NLP systems. In this study, we developed a novel prompt-based clinical NLP framework called HealthPrompt and applied the paradigm of prompt-based learning on clinical texts. In this technique, rather than fine-tuning or re-training a PLM, the task definitions are tuned by defining a prompt template. We find that, by carefully designing the prompt templates, PLMs can be effectively used for clinical text classification tasks. We performed an in-depth analysis of HealthPrompt on six different pre-trained language models in a no-data setting. Our experiments indicate that prompts could effectively capture the context of clinical texts in a zero-shot setting and perform well on clinical text classification without any training data. To the best of our knowledge, prompt-based ZSL has not been effectively applied to clinical NLP tasks, and hence the proposed HealthPrompt framework may be viewed as the foundation for a new paradigm for Clinical NLP, especially for NLU and NLI tasks.

**Acknowledgment**

This project was partially supported by the University of Pittsburgh Momentum Seed Funds and the National Institutes of Health through Grant Number UL1TR001857 funds. The funders had no role in the design of the study, collection, analysis, and interpretation of data and in preparation of the manuscript. The views presented in this report are not necessarily representative of the funder's views and belong solely to the authors.